\documentclass{article}
\usepackage{spconf,amsmath,graphicx}
\usepackage{algorithm,algpseudocode}
\usepackage{mathtools}
\usepackage{cancel}
\usepackage{amssymb}
\usepackage{booktabs}
\usepackage{caption}
\usepackage{subcaption}
\usepackage{multirow}
\usepackage{setspace}
\usepackage{icomma}
\usepackage{enumitem}
\usepackage{paralist}
\usepackage{txfonts}
\usepackage{pifont}
\usepackage{tikz}
\usetikzlibrary{positioning}
\usetikzlibrary{calc}

\usepackage[
backend=biber,
style=ieee,
citestyle=numeric-comp,
maxbibnames=3,
maxcitenames=3,
doi=false,isbn=false,url=false,eprint=false
]{biblatex}

\defbibheading{bibliography}[\refname]{}

\newcommand{\xmark}{\ding{55}}%

\DeclareSourcemap{
	\maps[datatype=bibtex, overwrite=true]{
		\map{
			\step[fieldsource=booktitle,
			match=\regexp{.*Interspeech.*},
			replace={Proc. Interspeech}]
			\step[fieldsource=journal,
			match=\regexp{.*INTERSPEECH.*},
			replace={Proc. Interspeech}]
			\step[fieldsource=booktitle,
			match=\regexp{.*ICASSP.*},
			replace={Proc. ICASSP}]
			\step[fieldsource=booktitle,
			match=\regexp{.*icassp_inpress.*},
			replace={Proc. ICASSP (in press)}]
			\step[fieldsource=booktitle,
			match=\regexp{.*Acoustics,.*Speech.*and.*Signal.*Processing.*},
			replace={Proc. ICASSP}]
			\step[fieldsource=booktitle,
			match=\regexp{.*International.*Conference.*on.*Learning.*Representations.*},
			replace={Proc. ICLR}]
			\step[fieldsource=booktitle,
			match=\regexp{.*International.*Conference.*on.*Computational.*Linguistics.*},
			replace={Proc. COLING}]
			\step[fieldsource=booktitle,
			match=\regexp{.*SIGdial.*Meeting.*on.*Discourse.*and.*Dialogue.*},
			replace={Proc. SIGDIAL}]
			\step[fieldsource=booktitle,
			match=\regexp{.*International.*Conference.*on.*Machine.*Learning.*},
			replace={Proc. ICML}]
			\step[fieldsource=booktitle,
			match=\regexp{.*North.*American.*Chapter.*of.*the.*Association.*for.*Computational.*Linguistics:.*Human.*Language.*Technologies.*},
			replace={Proc. NAACL}]
			\step[fieldsource=booktitle,
			match=\regexp{.*Empirical.*Methods.*in.*Natural.*Language.*Processing.*},
			replace={Proc. EMNLP}]
			\step[fieldsource=booktitle,
			match=\regexp{.*Association.*for.*Computational.*Linguistics.*},
			replace={Proc. ACL}]
			\step[fieldsource=booktitle,
			match=\regexp{.*Automatic.*Speech.*Recognition.*and.*Understanding.*},
			replace={Proc. ASRU}]
			\step[fieldsource=booktitle,
			match=\regexp{.*Spoken.*Language.*Technology.*},
			replace={Proc. SLT}]
			\step[fieldsource=booktitle,
			match=\regexp{.*Speech.*Synthesis.*Workshop.*},
			replace={Proc. SSW}]
			\step[fieldsource=booktitle,
			match=\regexp{.*workshop.*on.*speech.*synthesis.*},
			replace={Proc. SSW}]
			\step[fieldsource=booktitle,
			match=\regexp{.*Advances.*in.*neural.*information.*processing.*},
			replace={Proc. NeurIPS}]
			\step[fieldsource=booktitle,
			match=\regexp{.*Advances.*in.*Neural.*Information.*Processing.*},
			replace={Proc. NeurIPS}]
			\step[fieldsource=booktitle,
			match=\regexp{.*Workshop.*on.* Applications.* of.* Signal.*Processing.*to.*Audio.*and.*Acoustics.*},
			replace={Proc. WASPAA}]
			\step[fieldsource=publisher,
			match=\regexp{.+},
			replace={{}}]
			\step[fieldsource=month,
			match=\regexp{.+},
			replace={{}}]
			\step[fieldsource=location,
			match=\regexp{.+},
			replace={{}}]
			\step[fieldsource=address,
			match=\regexp{.+},
			replace={{}}]
			\step[fieldsource=organization,
			match=\regexp{.+},
			replace={{}}]
		}
	}
}
\title{Semi-autoregressive Streaming ASR with Label Context}
\name{Siddhant Arora${}^1$, George Saon${}^2$, Shinji Watanabe${}^1$, Brian Kingsbury${}^2$}
\address{${}^1$Carnegie Mellon University, ${}^2$IBM Research, Yorktown Heights, USA}
\begin{document}
\ninept
\maketitle
\begin{abstract}
Non-autoregressive (NAR) modeling has gained significant interest in speech processing since these models achieve dramatically lower inference time than autoregressive (AR) models while also achieving good transcription accuracy. Since NAR automatic speech recognition (ASR) models must wait for the completion of the entire utterance before processing, some works explore streaming NAR models based on blockwise attention for low-latency applications. However, streaming NAR models significantly lag in accuracy compared to streaming AR and non-streaming NAR models. To address this, we propose a \emph{streaming ``semi-autoregressive'' ASR model} that incorporates the labels emitted in previous blocks as additional context using a Language Model (LM) subnetwork. We also introduce a novel greedy decoding algorithm that addresses insertion and deletion errors near block boundaries while not significantly increasing the inference time. Experiments show that our method outperforms the existing streaming NAR model by 19\% relative on Tedlium2, 16\%/8\% on Librispeech-100 clean/other test sets, and 19\%/8\%  on the Switchboard(SWB)/Callhome(CH) test sets. It also reduced the accuracy gap with streaming AR and non-streaming NAR models while achieving 2.5x lower latency. We also demonstrate that our approach can effectively utilize \emph{external text data} to pre-train the LM subnetwork to further improve streaming ASR accuracy.

\end{abstract}
\begin{keywords}
ASR, Streaming, CTC, Semi-Autoregressive
\end{keywords}

\section{Introduction}
\label{sec:intro}
End-to-end Speech Recognition (E2E ASR) systems~\cite{prabhavalkar2023end,li2022recent,espnet,CTC,RNN-T,AED} have been widely studied thanks to their many applications like voice assistants and intelligent home devices. Most E2E architectures, like Recurrent Neural Network Transducer (RNN-T)~\cite{RNN-T,RNN-T2,RNN-T3} and Attention Encoder Decoder (AED)~\cite{AED,AED2,joint-ctc-att-mtl,joint-ctc-att-decoding,tuske2019advancing} systems, are Autoregressive (AR) models that condition on previous labels to make their predictions. While these AR models have achieved strong performance, their inference time increases with output length, thereby affecting user experience when they are deployed in interactive applications. To address this, prior works have introduced Non-Autoregressive (NAR) models~\cite{CTC,NAR1,NAR2,NAR3,NAR4,SelfCTC,InterCTC} that assume conditional independence on previously predicted labels and, hence, can output tokens concurrently. The Connectionist Temporal Classification (CTC)~\cite{CTC} loss is an NAR methodology that has been shown to achieve promising results while drastically reducing inference time. 

However, these models must wait for the utterance to end before they can start processing and hence cannot be deployed in low-latency applications. Further, these approaches are mostly based on the Transformer~\cite{transformer,conformer,conformer_another} architecture where the memory and computation requirement grows quadratically with the length of the input, making them impractical for handling very long utterances. Some attempts have been made to create framewise streaming ASR models using unidirectional encoders for RNN-T or CTC models; however, their performance is suboptimal~\cite{frame_streaming} due to no access to the future context. Hence, there has been interest in blockwise processing encoders~\cite{block_ASR1,block_asr2,block_asr3,block_asr4}. Prior work has tried to capture global information by introducing an additional context embedding vector~\cite{contextual_block} in each block. This approach paves the way for streaming ASR, achieved through blockwise synchronous inference~\cite{streaming_asr,streaming_asr_beam} of AED. Similar efforts~\cite{streaming_rnnt1,streaming_rnnt2,streaming_rnnt-3,streaming_rnnt4} have developed low-latency RNN-T ASR systems.

Inspired by the success of NAR models for offline ASR, there has been an effort to build streaming NAR~\cite{streaming_NAR} models that combines blockwise attention with CTC models and proposes a dynamic overlapping strategy during greedy decoding to address insertion and deletion errors at block boundaries. While streaming NAR models achieve dramatically lower latency, their accuracy is significantly worse than streaming AR and non-streaming NAR models. Prior work on machine translation has shown promising results with a semi-autoregressive (SAR)~\cite{semiAR_MT,semiAR_CV} model that retains the AR property globally but relaxes the AR property within local blocks. Motivated by this, we ask if we can encode labels predicted by previous blocks as additional context to improve streaming NAR accuracy without drastically increasing the inference time.

To this end, we propose a streaming SAR ASR model that performs greedy NAR decoding within a block but keeps the AR property across blocks by encoding the labels emitted at previous blocks using a Language Model (LM) subnetwork, effectively adding an additional ``label'' context vector in the contextual block encoder. This LM subnetwork is pretrained using a causal Language Modelling objective on text-only data, and hence provides \emph{text injection} functionality. During training, we employ teacher forcing and generate context embeddings using forced alignments~\cite{CTC_forced} obtained from CTC-based ASR models. Our entire model is trained using frame-wise cross-entropy loss, with the forced alignments serving as a proxy for ground truth alignments. We additionally experiment with intermediate CTC~\cite{InterCTC} and ``random block''~\cite{block-random,random_block2,random_block3,random_block4} regularisation. Finally, we apply a simple decoding strategy that combines the last few non-blank frames from the previous block with the frames in the current block during inference. We evaluated our approach on 3 publicly available datasets, Tedlium-2~\cite{tedlium2}, Librispeech-100~\cite{librispeech} and Switchboard~\cite{SWBD} . Our results show that our proposed streaming SAR model reduces the accuracy gap with streaming AR and non-streaming NAR models while achieving 2.5x lower latency.

The key contributions of our work are 
\begin{inparaenum}[(1)]
\item we introduce a novel streaming SAR ASR model that incorporates labels predicted from previous blocks as additional context,
\item we propose a novel decoding algorithm that improves over existing streaming NAR decoding strategies, and 
\item we show that our approach can pretrain the LM subnetwork with external text to boost streaming ASR accuracy.
\end{inparaenum}
\begin{figure}[t]
\centering
\includegraphics[width=0.5\textwidth]{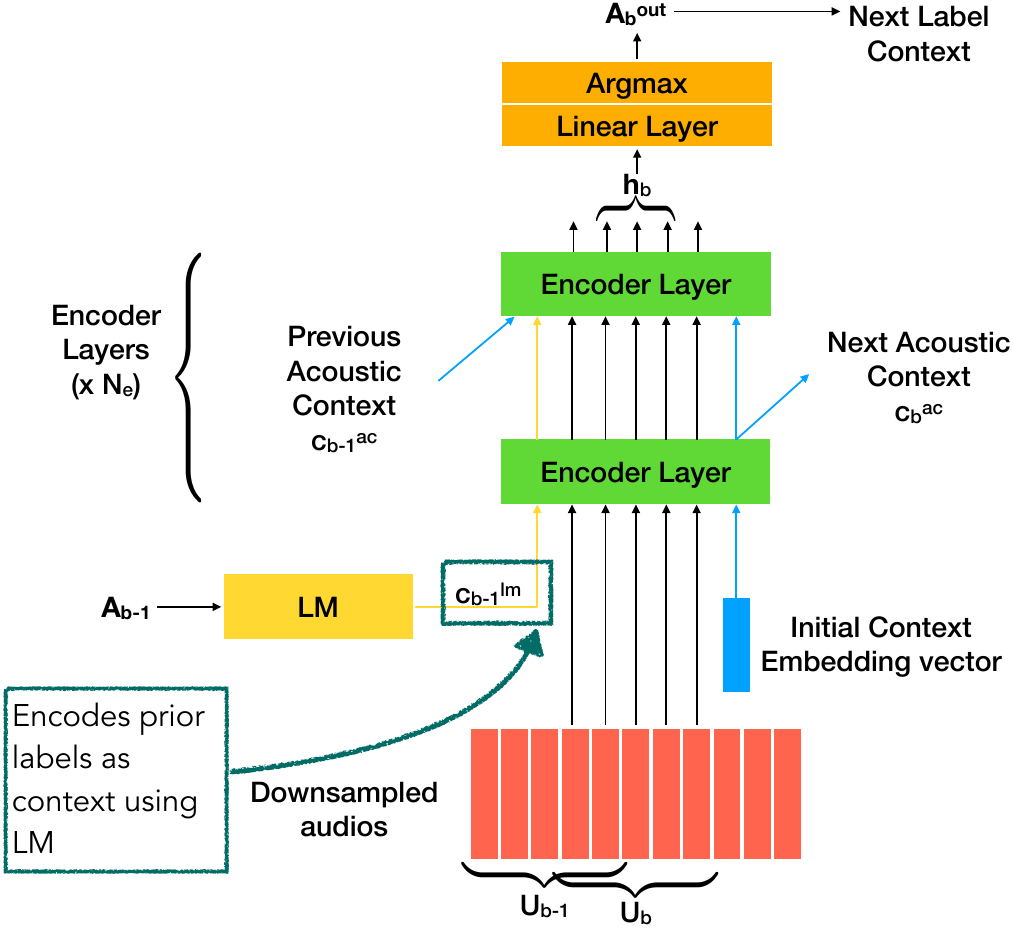}
\vskip -0.1in
\caption{Schematics of our semi-autoregressive streaming ASR model}
\label{fig:system-overview}
\vskip -0.2in
\end{figure}

\section{Problem Formulation}
\label{sec: problem formulation}
In ASR, the input is a $T$-length sequence of speech features, $X = \{\mathbf{x}_t | t=1, \dots, T\}$, and the goal is to predict the corresponding $O$-length transcript $Y = \{y_o| o=1, \dots, O\}$. 
Through the maximum a posteriori (MAP) decision theory, the ASR model estimates the transcript by maximizing posterior probability $P(Y|X)$.

In the streaming NAR model~\cite{streaming_NAR}, the input is represented by a sequence of $B$ overlapping blocks, $U = \{U_b \mid b=1, \dots, B\}$. Assuming a block size of $L_{\text{block}}$ and a hop size of $L_{\text{hop}}$, the $b^{th}$ block is $U_{b}=\mathbf{x}_{(I_{b-1}+1):(I_{b-1}+L_{block})}$ where $I_{b-1}=(b-1)*L_{\text{hop}}$.  Let $A = \{a_t \mid t=1, \dots, T\}$ be the \textit{frame-level alignment} where $a_t$ corresponds to the aligned transcript token for speech frame $\mathbf{x}_t$. 
Since the blocks overlap, streaming models~\cite{streaming_asr_beam} use the central $L_{\text{hop}}$ frames from each block for prediction, excluding the first $N_l$ frames as past frames and the last $N_r$ frames as future frames for lookahead.
We can approximate~\cite{CTC} posterior probability $P(Y|X) \approx \max_{A} P(A|X)$ and represent its logarithm as a sum of log-posteriors from overlapping blocks:
\begin{equation}
    \log(P(Y|X)) \approx \sum_{b=1}^{B} \sum_{t=I_{b-1}+N_l+1}^{I_{b-1}+N_l+L_{hop}} \log(P(a_{t}|a_{1:t-1},X))\label{eq:theory}.
\end{equation}
With the NAR~\cite{CTC} conditional independence (C.I.) assumption, $a_{t} \Perp a_{1:t-1} \mid X$, and the block processing~\cite{block_ASR1,block_asr2,block_asr3,block_asr4,contextual_block} C.I. assumptions, $a_{t} \Perp U_{<b} \mid U_{b}$\footnote{\cite{contextual_block} can potentially encode $U_{<b}$ using the acoustic context embedding} and $a_{t} \Perp U_{>b} \mid U_{b}$, Eq.~\ref{eq:theory} simplifies to
\begin{align}
\log(P(Y|X)) = \sum_{b=1}^{B} \sum_{t=I_{b-1}+N_l+1}^{I_{b-1}+N_l+L_{hop}} \log(P(a_{t}|U_{b})).
\label{eq:streaming_NAR}
\end{align}

We propose to preserve the AR property globally.
We relax the NAR C.I. assumption and modify Eq.~\ref{eq:streaming_NAR} by conditioning on labels emitted in previous blocks, $A_{b-1}=a_{1:(I_{b-1}+N_{l})}$, using LM subnetwork:
\begin{align}
\log(P(Y|X)) = \sum_{b=1}^{B} \sum_{t=I_{b-1}+N_l+1}^{I_{b-1}+N_l+L_{hop}} \log(P(a_{t}|\underset{\text{LM Sub-Net}}{\underbrace{A_{b-1}}},U_b)).
\label{eq:semi_auto_NAR}
\end{align}
By relaxing the NAR C.I. assumption, our formulation better approximates the original posterior distribution in Eq.~\ref{eq:theory}.
This formulation enhances the streaming NAR accuracy using the ``label context'' embedding from the LM subnetwork while maintaining low latency by being able to output tokens concurrently within a block.
\begin{table*}[t]
  \centering
    \resizebox {0.99\linewidth} {!} {
\begin{tabular}{clllcccccccc}
\toprule
& & & & \multicolumn{2}{c}{Tedlium-2} & \multicolumn{3}{c}{Librispeech-100} & \multicolumn{3}{c}{SWB}\\ 
\cmidrule(r){5-6}\cmidrule(r){7-9}\cmidrule(r){10-12}
\texttt{ID} & Model & Encoder Type & Decode Mode &  WER & Latency & Clean WER & Other WER & Latency & SWB WER & CH WER & Latency\\ 
 \midrule
 \texttt{A1} &  & \multirow{2}{*}{Transformer} & Full Path & 12.5 & 1102 & 12.2 & 30.1 & 1170 & \xmark & \xmark & \xmark\\
\texttt{A2} & Non Streaming NAR & & Greedy & 12.9 & \hphantom{0}391 & 12.3 & 30.3 &  \hphantom{0}408 & \xmark & \xmark & \xmark\\
  \cmidrule(r){3-12}
\texttt{A3} & (topline)& \multirow{2}{*}{Conformer} & Full Path & \hphantom{0}8.8 & 1806 & \hphantom{0}8.8 & 22.8 & 1531 & \hphantom{0}9.1 & 16.7 & 409 \\
\texttt{A4} & & & Greedy & \hphantom{0}8.9 & \hphantom{0}872 & \hphantom{0}8.8 & 22.9 & \hphantom{0}752 & \hphantom{0}9.1 & 16.5 & 270\\
 \midrule
 \texttt{B1} & \multirow{2}{*}{Streaming AR} & Transformer & Full Path & 11.5  & \hphantom{0}599 & 10.7 & 26.8 & \hphantom{0}540 & \xmark & \xmark & \xmark\\
\texttt{B2}  & & Conformer & Full Path & 10.5 & \hphantom{0}776 & \hphantom{0}8.9 & 26.1 & \hphantom{0}444 & \hphantom{0}9.3 & 16.8 & 411\\
 \midrule
 \texttt{C1} & \multirow{6}{*}{Streaming NAR} & \multirow{3}{*}{Transformer} & Full Path & 14.9 & \hphantom{0}327 & 12.5 & 32.7 & \hphantom{0}305 & \xmark & \xmark & \xmark\\
\texttt{C2} &  & & Greedy & 18.2 & \hphantom{00}46 & 16.6 & 35.1 & \hphantom{00}78 & \xmark & \xmark & \xmark\\
 \texttt{C3} & & & Overlap Greedy~\cite{streaming_NAR} & 16.0 & \hphantom{00}53 & 13.4 & 33.2 & \hphantom{0p}80 & \xmark & \xmark & \xmark\\
 \cmidrule(r){3-12}
\texttt{C4}  & & \multirow{3}{*}{Conformer} & Full Path & 11.0 & \hphantom{0}482 & \hphantom{0}9.4 & 27.0 & \hphantom{0}453 & 10.7 & 18.4 & 304 \\
\texttt{C5}  & & & Greedy & 15.1 & \hphantom{0}124 & 16.0 & 31.6 & \hphantom{0}146  & 14.6 & 20.3 & 115\\
 \texttt{C6} & & & Overlap Greedy & 12.5 & \hphantom{0}124 & 10.8 & 28.1& \hphantom{0}144& 12.4 & 19.1 & 116\\
 \midrule
 \texttt{D1} & Streaming NAR & \multirow{3}{*}{Conformer} & Full Path & 12.0 & \hphantom{0}487 & \hphantom{0}9.8 & 28.1 & \hphantom{0}454 & 10.5 & 17.7 & 303\\
 \texttt{D2} &  trained w/ alignment & & Greedy & 16.4 & \hphantom{0}121 & 17.0 & 33.2 & \hphantom{0}146 & 14.9 & 20.2 & 114\\
 \texttt{D3} & & & Overlap Greedy & 13.6 & \hphantom{0}123 & 12.1 & 29.8 & \hphantom{0}145 & 12.5 & 18.8 & 115\\
 \midrule
\texttt{E1} &  \multirow{4}{*}{Streaming SAR} & \multirow{2}{*}{Conformer} & Overlap Greedy & 11.7 & \hphantom{0}143 & 12.3 & 29.2 & \hphantom{0}188  & 12.4 & 18.8 & 133\\
 \texttt{E2} & & & Alignment Greedy & 10.5 & \hphantom{0}140 & \hphantom{0}9.6 & 26.9 & \hphantom{0}188 & 10.4 & 17.7 & 137\\
 \texttt{E3} & & \hphantom{00}+InterCTC & Alignment Greedy & 10.2 & \hphantom{0}144 & \hphantom{0}9.6 & 26.6 & \hphantom{0}193 & 10.3 & 17.6 & 136\\
 \texttt{E4} & & \hphantom{00}\hphantom{00}+Random block & Alignment Greedy & 10.1 & \hphantom{0}149 & \hphantom{0}9.1 & 26.1 & \hphantom{0}188  & 10.1  & 17.5 &  136 \\
 
\bottomrule
\end{tabular}
}
\vskip -0.13in
\caption{Results presenting ASR accuracy and latency (in msec.) of our semi-autoregressive (SAR) ASR system on Tedlium-2, Librispeech-100 and Switchboard dataset. Due to time constraints, we opted to report the performance of only conformer-based models for Switchboard, given their superior performance over transformer-based models on Librispeech-100 and Tedlium-2 datasets.} 
\vskip -0.18in
\label{tab:main-results}
\end{table*}
\section{Method}
\label{sec: method}
To achieve the formulation described in Eq.~\ref{eq:semi_auto_NAR}, we propose the streaming SAR ASR model shown in Figure~\ref{fig:system-overview}. For the $b^{th}$ block, the output of the previous blocks $A_{b-1}$ are passed through an LM subnetwork to produce the label context embedding $\textbf{c}^{\text{lm}}_{b-1}$:
\begin{equation}
    \textbf{c}^{\text{lm}}_{b-1}= \text{LM}(\text{Normalize}(A_{b-1}))\label{lm_eq},
\end{equation}
where ``Normalize'' refers to removing repeated and blank tokens from alignments to make it similar to the transcripts used for training LMs such that $\text{Normalize}(A)=Y$.

The input speech features $U_b$ for each block $b$ are passed to the contextual block encoder (CBE) along with the context embeddings:
\begin{equation}
    \textbf{h}_{b}, \textbf{c}^{\text{ac}}_{b}= \text{CBE}(U_b, \textbf{c}^{\text{lm}}_{b-1}, \textbf{c}^{\text{ac}}_{b-1})\label{aco_eq}
\end{equation}
where $\textbf{c}^{\text{ac}}_{b-1}$ is the previous acoustic context embedding and $\textbf{c}^{\text{ac}}_{b}$ is the next acoustic context embedding~\cite{contextual_block}.
To train the model, we cannot employ an alignment-free CTC loss since we need to ensure that the frame-level output of our model aligns with the labels used for computing label context embeddings $A_{b-1}$ in Eq.~\ref{lm_eq}. This is crucial to prevent a train-test mismatch since we train using teacher forcing of alignments, while during inference the model utilizes labels produced in previous blocks. Consequently, we train our entire model utilizing frame-level cross-entropy loss, denoted as $\text{Loss}_{\text{CE}}$:
\begin{equation}
    \text{Loss} = \text{Loss}_{\text{CE}}(\text{Softmax}(\text{Out}(\cup_{b=1}^{B}\textbf{h}_{b})), A)\label{cross_entropy_eq}
\end{equation}
where $\text{Out}(\cdot)$ denotes a linear layer that maps encoder output $\textbf{h}_{b}$ to the vocabulary size $|\mathcal{V}|$ followed by a softmax function. 

\subsection{Training and Inference}
\label{subsec: training}

\begin{algorithm}
\caption{Alignment greedy decoding}
\begin{algorithmic}[1]
    \State $Y^{\text{out}}$ = [$\varnothing$]\textbf{};
    \State $A_0$ = [$\varnothing$]\textbf{};
    \State $Y^{\text{prev}}$ = [$\varnothing$]\textbf{};
    \State $\mathbf{U}$ = Contextual Audio blocks iterators;
    \For{$b$ = 0 to B}\;
    \State $\textbf{c}^{\text{lm}}_{b-1} =$LM(\text{Normalize}($A_{b-1}$))\;
    \State $A^{out}_{b}$ = Argmax(Out(CBE($U_b$,$\textbf{c}^{\text{lm}}_{b-1}$)))\;
    \State $A_b \leftarrow \{A_{b-1}, A^{out}_{b}\}$\;
    \State $A^{out}_{b} \leftarrow \{Y^{\text{prev}}, A^{out}_{b}\}$\;
    \If{$ A^{out}_{b} $ ends in non-blank token and is not final}
    \State $Y^{\text{prev}}$ = frames with ending token in $A^{out}_{b}$;
    \State $A^{out}_{b}$ = remove frames with ending token in $A^{out}_{b}$;
    \Else
    \State $Y^{\text{prev}}$ = [$\varnothing$]\textbf{};
    \EndIf 
    \State $Y_{b}$ = \text{Normalize}($A^{out}_{b}$)\;
    \State $Y^{\text{out}} \leftarrow \{Y^{\text{out}}, Y_{b}\}$\;
    \EndFor
    \State \Return $Y^{\text{out}}$
\end{algorithmic}
\vspace{-3pt}
\end{algorithm}

We use CTC forced alignments~\cite{CTC_forced} from a pre-trained NAR ASR model as a proxy to obtain frame-level alignment $A$ for training.  
We use $A$ both for computing label context embeddings (see Eq.~\ref{lm_eq}) and as ground truth for Cross Entropy Loss (see Eq.~\ref{cross_entropy_eq}).  
Further, we pre-train the LM subnetwork using a causal LM objective on training transcripts $Y$. This formulation further allows us to \emph{incorporate external text data} to pre-train the LM subnetwork. 

During inference, we use the frame level output from the previous blocks as $A_{b-1}$ to produce label context embeddings using Eq.~\ref{lm_eq}. Prior works~\cite{streaming_NAR} have shown that the straightforward approach of splitting input audio into smaller blocks and performing greedy decoding on each block results in significant accuracy degradation. We similarly observed that most ASR errors occur when the segment boundary appears in the middle of a token, leading to repetitive recognition in 2 consecutive blocks. Based on this observation, we came up with a simple decoding strategy outlined in Algorithm 1, which we refer to as ``alignment greedy decoding". Let the frame-level output from each block be 
denoted by $A^{out}_b$, then the 
frame-level alignment $A_{b-1}$ is first updated to $A_{b}$ by appending it with $A^{out}_b$. If $A^{out}_b$ ends with non-blank frames, this decoding removes the last non-blank frames $Y^{\text{prev}}$ from the current block $A^{out}_b$ and instead adds $Y^{\text{prev}}$ to the next block $A^{out}_{b+1}$.

\section{Experiments}

\subsection{Datasets}
To demonstrate the effectiveness of our streaming SAR model (see section~\ref{sec: method}), we conducted experiments on 3 publicly available English ASR datasets: Tedlium-2~\cite{tedlium2}, Librispeech-100~\cite{librispeech} and Switchboard (SWB)~\cite{SWBD} . 
We quantify ASR accuracy using Word Error Rate (WER). For non-streaming models, latency is straightforwardly represented by the average decoding time per utterance.
For streaming models, we calculate latency as $\text{Latency}=\frac{\sum_{utt}(\text{last token emitted}-\text{duration of speech})}{\#\text{total number of utterance}}
$.
Here, ``last token emitted'' denotes the moment when the model emits the final token, while ``duration of speech'' refers to the entire duration of the input audio. When computing the timestamp at which the model outputs the last token, we take into account both the lookahead time, which encompasses the time required before processing the next block can start, and the actual processing time.

\subsection{Baseline}
We compare our proposed streaming SAR model with streaming NAR and AR models.  
In addition, we include a non-streaming NAR model as a topline, demonstrating how our approach can effectively narrow the accuracy gap while maintaining significantly lower latency.
Furthermore, we present the results of a streaming NAR model trained with frame-level cross-entropy loss (Eq.~\ref{cross_entropy_eq}) using alignments to gain insights into the impact of alignment-based training. We also incorporate intermediate CTC~\cite{lee2021intermediate} and ``random block'' regularization~\cite{block-random} techniques into our streaming SAR system.

\subsection{Experimental Setups}
For our NAR models, we employ a 12-layer transformer with a hidden dimension of 256 for Librispeech-100 and Tedlium2 datasets. Our conformer NAR model comprises 12 layers with a hidden dimension of 256 for Tedlium2 and SWB, while for Librispeech-100, it consists of 18 layers with a hidden dimension of 256. The streaming AR models are constructed with a contextual block encoder using $L_{\text{block}}=40$, $L_{\text{hop}}=16$, $N_l=8$, $N_r=16$ and share the same hyperparameters as the NAR models. The decoder for the streaming AR models is a 6 layer transformer. 
Streaming NAR baselines consist of the same hyperparameters as the streaming AR encoders.
Our streaming SAR model shares the same blockwise encoder architecture as the streaming NAR model and incorporates an LM subnetwork. The LM subnetwork consists of an LSTM with 2 layers and a hidden dimension of 256 for Tedlium2, and 1 layer with a hidden dimension of 1024 for Librispeech-100 and SWB. We experiment with the incorporation of intermediate CTC~\cite{lee2021intermediate} after the 6th layer with weight of 0.3. We also explore the effectiveness of ``random block'' regularization~\cite{block-random}, where we randomly select block sizes from the range of [35, 45] during training, instead of using a fixed block size. We use bpe size of 500 for all datasets. 

The streaming AR models are trained using joint CTC-attention training~\cite{joint-ctc-att-mtl}, while the streaming and non-streaming NAR models are trained using the CTC~\cite{CTC} loss.
We train a streaming NAR model and our proposed streaming SAR models with cross-entropy loss (Eq.~\ref{cross_entropy_eq})  using frame-level alignments. As ground truth alignments are unavailable, we utilize forced alignments from the streaming conformer NAR model as a proxy. We train the LM subnetwork on training transcripts. We also experiment with training on external text data\footnote{\url{http://www.openslr.org/resources/11/librispeech-lm-norm.txt.gz}},
including the transcripts of the entire Librispeech for the Librispeech-100 and the Fisher transcripts for the SWB SAR model. 

 We use both greedy and full-path decoding for the non-streaming NAR model. We run full-path decoding for the streaming AR and streaming NAR models via blockwise synchronous inference~\cite{streaming_asr_beam,streaming_asr}. Additionally, we employ greedy decoding along with dynamic mapping for overlapping inference~\cite{streaming_NAR} (referred to as ``overlap decoding'') for our streaming NAR models.
We perform inference for streaming SAR model using both ``overlap decoding'' and our proposed ``alignment decoding'' (see Algorithm 1). 
The inference of our models are performed using 4 parallel CPU (AMD EPYC 7763 @ 2.55 GHz) jobs with 64 GB memory. All model, training and inference parameters are selected based on validation accuracy.\footnote{Full details regarding models, configuration files, and data preparation setup will be made publicly available as part of the ESPnet~\cite{espnet} toolkit.}
\begin{table}[t]
  \centering
 \resizebox {0.65\linewidth} {!} {
\begin{tabular}{lc}
\toprule
Model & WER\\ 
 \midrule
 Streaming NAR trained w/ alignment & \\
 \hphantom{000} streaming NAR alignment & 13.6 \\
 \hphantom{000} non streaming NAR alignment & 14.6 \\
 \midrule
 Streaming SAR & \\
 \hphantom{000}LM subnetwork & 11.7 \\
 \hphantom{000}\hphantom{000} w/o finetuning & 12.0 \\
 \hphantom{000}\hphantom{000} w/o LM pre-training & 12.4 \\
 \hphantom{000}Cross Attention & 13.6 \\
\bottomrule
\end{tabular}
}
\vskip -0.13in
\caption{Results presenting ablation on different ways of using alignment to incorporate label context. Results are shown for conformer based models with overlap decoding on Tedlium 2 dataset.} 
\vskip -0.2in
\label{tab:ablation-results1}
\end{table}

\subsection{Results and Discussion}
\label{subsec: result}
Table~\ref{tab:main-results} presents the performance of our streaming SAR models alongside topline and baseline models. We observe that the streaming NAR model with greedy (\texttt{C5}) and overlap greedy (\texttt{C6}) decoding~\cite{streaming_NAR} achieves impressively low latency but significantly degrades accuracy compared to streaming AR (\texttt{B2}) and non-streaming NAR (\texttt{A4}) models.
Our experiments reveal that training with alignments using cross-entropy loss mostly yield inferior results compared to training with CTC loss (\texttt{D3} vs \texttt{C6}), possibly due to imperfections in frame-level alignments. Notably, our streaming SAR model improves accuracy (\texttt{E1} vs \texttt{D3}) by incorporating the LM subnetwork to encode the label context. 
Additionally, our proposed ``alignment decoding'' proves to be a valuable enhancement, delivering a significant boost in accuracy (\texttt{E2} vs \texttt{E1}) compared to the ``overlap decoding'' proposed in prior work~\cite{streaming_NAR}.
Further improvements are achieved by introducing intermediate CTC (\texttt{E3}) and employing ``random block'' (\texttt{E4}) regularization. Our best-performing streaming SAR model (\texttt{E4}) surpasses the streaming NAR model with overlap decoding (\texttt{C6}) by a relative margin of 19\% on Tedlium-2, 16\%/8\% on the clean/other test sets of Librispeech-100, 19\%/8\%  on the Switchboard(SWB)/Callhome(CH) test sets of SWB with only a slight increase in latency. We further conduct significance tests and observe a p-value of less than 0.003 using Matched Pair, Signed Paired, Wilcoxon and McNemar tests for all test datasets.
Moreover, our streaming SAR model outperforms the streaming NAR model with full path decoding (\texttt{E4} vs \texttt{C4}) by a relative 8\%, 4\%, 5\% on Tedlium-2, Librispeech-100 and SWB, respectively, while achieving a 2.5x reduction in latency. Impressively, our proposed streaming SAR model effectively bridges the accuracy gap with non-streaming NAR models (\texttt{E4,C6} vs \texttt{A4}) and can even match or surpass streaming AR models (\texttt{E4} vs \texttt{B2}) on Tedlium2 and Librispeech-100 all while delivering a 5x and 2.5x faster processing speed respectively.
\begin{table}[t]
  \centering
 \resizebox {\linewidth} {!} {
\begin{tabular}{lccccc}
\toprule
&  \multicolumn{2}{c}{Tedlium-2} & \multicolumn{3}{c}{Librispeech-100}\\ 
\cmidrule(r){2-3}\cmidrule(r){4-6}
LSTM-LM & Perplexity & WER & Perplexity & Clean WER & Other WER\\ 
 \midrule
2 layer, Dim 256 & 28.9 & 10.5 & 29.9 & 10.2 & 27.6\\
1 layer, Dim 1024 & 21.5 & 10.7 & 26.1 & 10.0 & 27.4 \\
\hphantom{000} w/ external data & - & - & 16.4 & \hphantom{0}9.6 & 26.9\\

\bottomrule
\end{tabular}
}
\vskip -0.13 in
\caption{Results presenting ablation on the impact of causal LM pre-training. Results are shown for the conformer based streaming SAR model with alignment decoding on Tedlium 2 and Librispeech-100 dataset along with perplexity of pre-trained LMs.} 
\vskip -0.2 in
\label{tab:ablation-results2}
\end{table}

\textbf{Ablation study on incorporating label context}: Table~\ref{tab:ablation-results1} shows our findings for conformer based models with ``overlap decoding'' inference on Tedlium2 dataset. First, we experiment with using ``forced alignment'' from a non-streaming NAR model and observe that forced alignments from a streaming NAR model leads to better accuracy. In the case of the LM subnetwork, we explore not fine-tuning the LM during training of the streaming SAR model, which proves that proposed finetuning is effective. We also experiment with not pre-training the subnetwork on a causal LM objective, and again observe that our formulation of initialising the LM subnetwork is helpful. Additionally, we conduct an ablation study where, instead of using an LM subnetwork, we experiment with incorporating previously predicted labels using the multi-sequence cross-attention~\cite{Compositional_E2E,cross-attention} in the decoder and observe that our proposed formulation of having a LM subnetwork achieves better accuracy. 

\textbf{Ablation study on impact of causal LM pre-training}: 
Table~\ref{tab:ablation-results2} shows the perplexity of different pre-trained language models (LMs) and their impact on streaming SAR accuracy. Our observations reveal that, for Tedlium2, there is no significant accuracy difference associated with a stronger LM. However, in the case of Librispeech-100, using a slightly superior LM results in an accuracy boost. Notably, the use of external text data further enhances accuracy, underscoring the effectiveness of our approach.

\section{Conclusion}
\label{sec: conclusion}
We introduce a novel streaming SAR model that performs NAR decoding within a block while maintaining the AR property across blocks by encoding labels emitted in previous blocks using an LM subnetwork. Our approach includes a simple alignment decoding algorithm that helps to mitigate recognition errors due to block boundaries. We demonstrate the effectiveness of pre-training the LM subnetwork with external text data. We also incorporate intermediate CTC and ``random block'' regularization. Our experiments reveal superior accuracy compared to streaming NAR models and also competitive accuracy with streaming AR models while achieving a 2.5x reduction in latency. 
Future work will explore the use of external text data via advanced LM~\cite{llama2} or CTC-based text injection~\cite{adaptation_interctc}.

\section{Acknowledgements}
This work used NCSA Delta through allocation CIS210014 from the Advanced Cyberinfrastructure Coordination Ecosystem: Services \& Support (ACCESS) program, which is supported by NSF grants \#2138259, \#2138286, \#2138307, \#2137603, \#2138296.

\vfill\pagebreak

\section{References}
{
\sloppy
\printbibliography
}

\end{document}